\documentclass[final]{cvpr}

\usepackage{times}
\usepackage{epsfig}
\usepackage{graphicx}
\usepackage{amsmath}
\usepackage{amssymb}

\usepackage{booktabs}
\usepackage{multirow}
\usepackage[usenames,dvipsnames]{color}
\usepackage{colortbl}

\usepackage{pifont}
\newcommand{\cmark}{\ding{51}}
\newcommand{\xmark}{\ding{55}}

\definecolor{mygray}{gray}{.9}
\definecolor{mypink}{rgb}{.99,.91,.95}
\definecolor{mycyan}{cmyk}{.3,0,0,0}
\newcolumntype{g}{>{\columncolor{mygray}}c}
\newcolumntype{w}{>{\columncolor{mygray}}l}

\newcommand*{\affaddr}[1]{#1} 
\newcommand*{\affmark}[1][*]{\textsuperscript{#1}}
\newcommand*{\email}[1]{\texttt{#1}}

\usepackage[pagebackref=true,breaklinks=true,colorlinks,bookmarks=false]{hyperref}

\def\cvprPaperID{4163} 
\def\confYear{CVPR 2021}

\begin{document}

\title{FSCE: Few-Shot Object Detection via Contrastive Proposal Encoding}

\author{Bo Sun\affmark[1], Banghuai Li\affmark[2]\thanks{Corresponding author: libanghuai@megvii.com}, Shengcai Cai\affmark[2], Ye Yuan\affmark[2], and Chi Zhang\affmark[2]\\
\affaddr{\affmark[1]University of Southern California}\\
\affaddr{\affmark[2]MEGVII Technology}\\
\email{bos@usc.edu, \{libanghuai,caishengcai,yuanye,zhangchi\}@megvii.com}\\
}

\maketitle

\begin{abstract}
\vspace{-2.5mm}
\ \ \ \ Emerging interests have been brought to recognize previously unseen objects given very few training examples, known as few-shot object detection (FSOD). Recent researches demonstrate that good feature embedding is the key to reach favorable few-shot learning performance.  We observe object proposals with different Intersection-of-Union (IoU) scores are analogous to the intra-image augmentation used in contrastive approaches. And we exploit this analogy and incorporate supervised contrastive learning to achieve more robust objects representations in FSOD. We present \textbf{F}ew-\textbf{S}hot object detection via \textbf{C}ontrastive proposals \textbf{E}ncoding (\textbf{FSCE}), a simple yet effective approach to learning contrastive-aware object proposal encodings that facilitate the classification of detected objects. We notice the degradation of average precision (AP) for rare objects mainly comes from misclassifying novel instances as confusable classes. And we ease the misclassification issues by promoting instance level intra-class compactness and inter-class variance via our contrastive proposal encoding loss (CPE loss). Our design outperforms current state-of-the-art works in any shot and all data splits, with up to $+8.8\%$ on standard benchmark PASCAL VOC and $+2.7\%$ on challenging COCO benchmark. Code is available at: \url{https://github.com/MegviiDetection/FSCE}.
\end{abstract}

\begin{figure}[h]
\begin{center}
\includegraphics[width=0.98\linewidth]{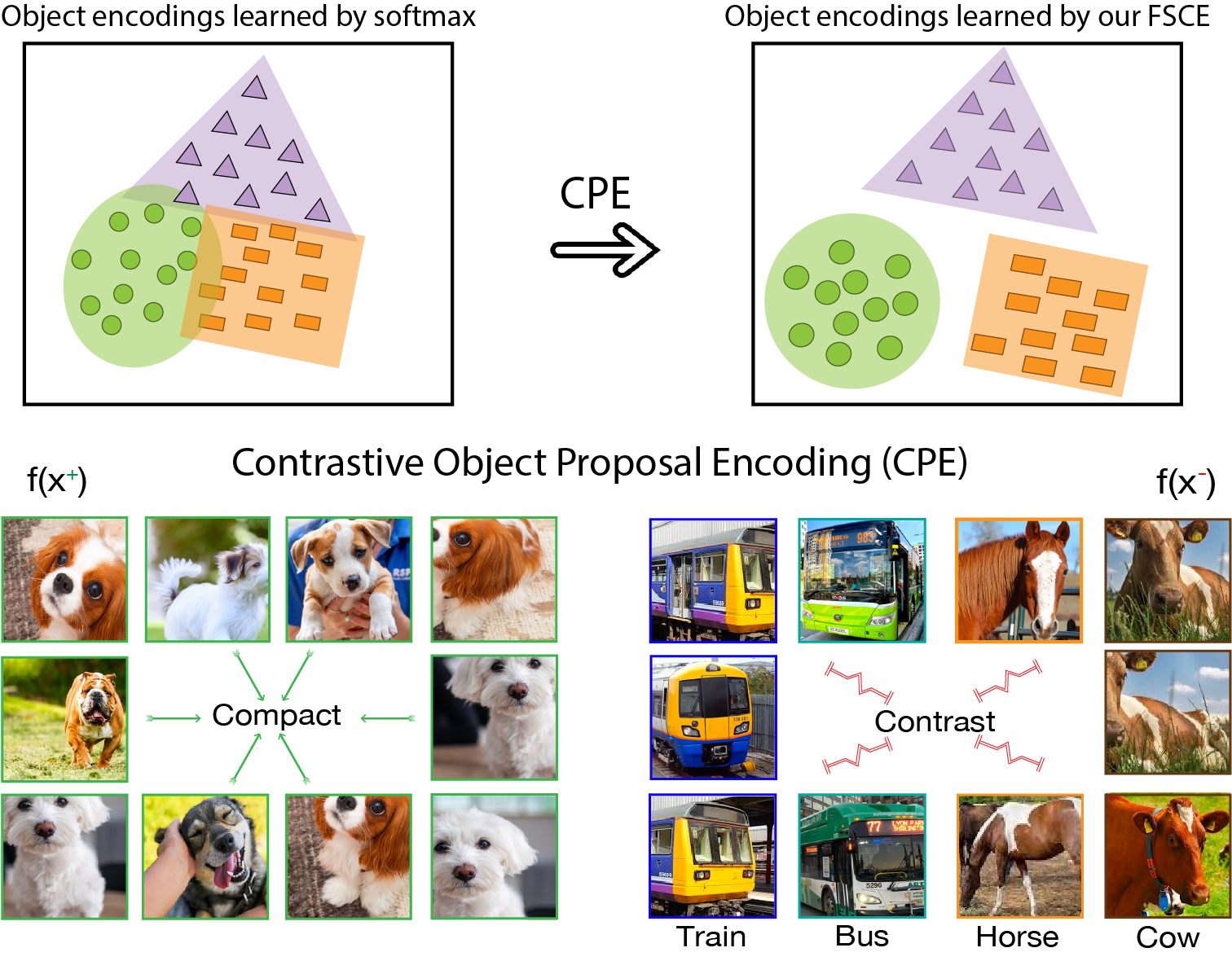}
\caption{Conceptualization of our contrastive object proposals encoding. We introduce a score function which measures the semantic similarity between region proposals. Positive proposals ($x^+$) refer to region proposals from the same category or the same object. Negative proposals ($x^-$) refer to proposals from different categories. We encourage the object encodings to have the property that $score(f(x),f(x^+))~\textgreater\textgreater~score(f(x),f(x^-))$, such that our contrastively learned object proposals have smaller intra-class variance and larger inter-class difference}
\vspace{-3mm}
\label{fig:figure1} 
\end{center}
\end{figure}

\begin{figure*}[t]
\includegraphics[width=1\textwidth]{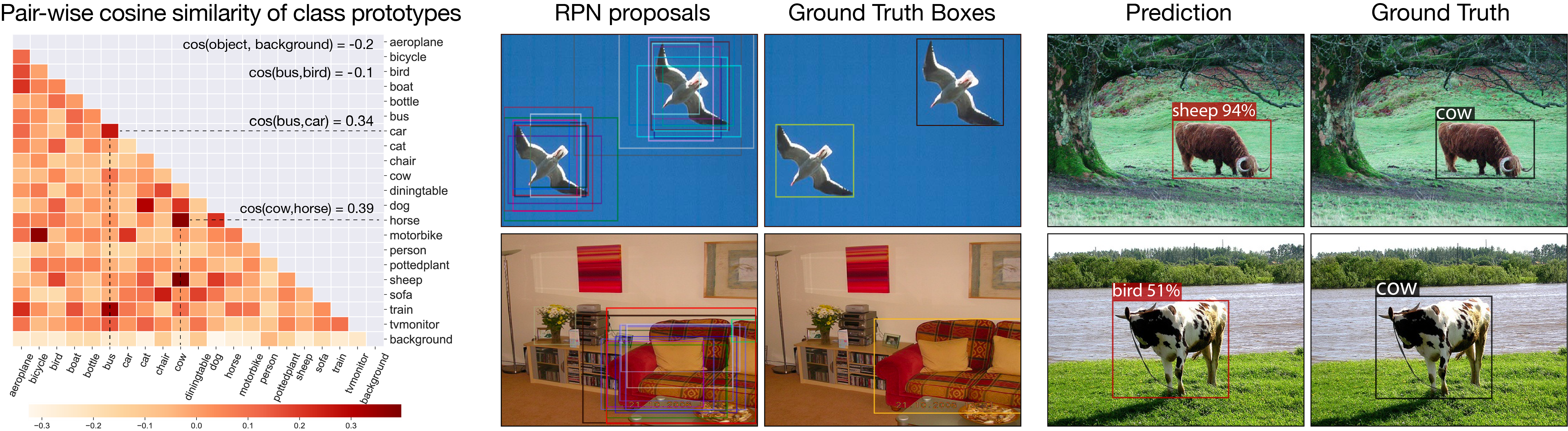}
\centering
\caption{We find in fine-tuning based few-shot object detector, classification is more error-prone than localization. In the fine-tuning stage, RPN is able to make good enough foreground proposals for novel instances, hence novel objects are often accurately localized but mis-classified as confusable base classes. Here shows 20 top-scoring RPN proposals and example detection results from PASCAL VOC Split 1, wherein \textit{bird}, \textit{sofa} and \textit{cow} are novel categories. The left panel shows the pair-wise cosine similarity between the class prototypes learned in the bounding box classifier. For example, the similarity between \textit{bus} and \textit{bird} is -0.10, but the similarity between \textit{cow} and \textit{horse} is 0.39. Our goal is to decrease the instance-level similarity between \textbf{similar} objects that are from \textbf{different} categories.}
\label{fig:pw-sim}
\end{figure*}

\vspace{-4mm}
\section{Introduction}
Development of modern convolutional neural networks (CNNs) \cite{he_deep_2015,cai_cascade_2017,zhu_deformable_2018} give rise to great advances in general object detection \cite{ren_faster_2016,lin_focal_2018,tian_fcos_2019}. Deep detectors demand a large amount of annotated training data to saturate its performance~\cite{tan_efficientdet_2020,wang_cspnet_2019}. In few-shot learning scenarios, deep detectors suffer severer over-fitting and the gap between few-shot detection and general object detection is larger than the corresponding gap in few-shot image classification~\cite{wang_low-shot_2018,khodadadeh_unsupervised_2019,ren_meta-learning_2018}. On the contrary, a child can rapidly comprehend new visual concepts and recognize objects from a newly learned category given very few examples. Closing such gap is therefore an important step towards more successful machine perception \cite{funke_five_2020}. 

Precedented by few-shot image classification, earlier attempts in few-shot object detection utilize meta-learning strategy \cite{yan_meta_2020,wang_meta-learning_2019,xiao_few_shot_2020}. Meta-learners are trained with an episode of individual tasks, meta-task samples from common objects (base class) to pair with rare objects (novel class) to simulate few-shot detection tasks. Recently, the two-stage fine-tune based approach (TFA) reveals more potential in improving few-shot detection. Baseline TFA \cite{wang_frustratingly_2020} simply freeze all base class trained parameters and fine-tune only box classifier and box regressor with novel data, yet outperforms previous meta-learners. MPSR \cite{wu_multi-scale_2020} improves upon TFA by alleviating the scale bias inherent to few-shot dataset, but their positive refinement branch demands manual selection, which is somewhat less neat. In this work, we observe and address the essential weakness of the fine-tuning based approach~--~constantly mislabeling novel instances as confusable categories, and improve the few-shot detection performance to the new state-of-the-art (SOTA).

Object detection involves localization and classification of appeared objects. In few-shot detection, one might naturally conjecture the localization of novel objects is going to under-perform its base categories counterpart, with the concern that rare objects would be deemed as background~\cite{wang_meta-learning_2019,yan_meta_2020,fan2020few}. However, based on our experiments with Faster R-CNN~\cite{ren_faster_2016}, the commonly adopted detector in few-shot detection, class-agonistic region proposal network (RPN) is able to make foreground proposals for novel instances, and the final box regressor can localize novel instances quite accurately. In comparison, as demonstrated in Figure~\ref{fig:pw-sim}, misclassifying detected novel instances as confusable base classes is indeed the main source of error. We visualize the pairwise cosine similarity between class prototypes~\cite{snell_prototypical_2017,cos_face,deng_arcface_2019} of a Faster R-CNN box classifier trained with PASCAL VOC~\cite{voc07,voc12}. The cosine similarity between prototypes from resembled categories can be $0.39$, whereas the similarity between objects and background is on average $-0.21$. In few-shot setting, the similarity between cluster centers can go as high as 0.59, \eg, between \textit{sheep} and \textit{cow}, \textit{bicycle} and \textit{motorbike}, making classification for similar objects error-prone.  We make a calculation upon baseline TFA, manually correcting misclassified yet accurately localized box predictions can increase novel class average precision (nAP) by over 20 points.

A common approach to learn well-separated decision boundary is to use a large margin classifier~\cite{large_margin}, but with our trials, category-level positive-margin based classifiers does not work in this data-hunger setting~~\cite{cos_face,deng2009imagenet}. To learn instance-level discriminative feature representations, contrastive learning~\cite{hadsell2006dimensionality,1467314} has demonstrated its effectiveness in tasks including recognition~\cite{schroff2015facenet}, identification~\cite{sun2014deep} and the recent successful self-supervised models~\cite{wu_unsupervised_2018,xie2020delving,he_momentum_2020,chen_simple_2020}. In supervised contrastive learning for image classification~\cite{supervised_contrastive_learning}, intra-image augmentations of images from the same class are used to enrich the positive example pairs. We think region proposals with different Intersection-over-Union (IoU) for an object are naturally analogous to the intra-image augmentation \textit{cropping}, as illustrated in Figure ~\ref{fig:figure1}. Therefore in this work, we explore to extend the supervised batch contrastive approach~\cite{supervised_contrastive_learning} to few-shot object detection. We believe the contrastively learned object representations aware of the intra-class compactness and the inter-class difference can ease the misclassification of unseen objects as similar categories.

We present \textbf{F}ew-\textbf{S}hot object detection via \textbf{C}ontrastive proposals \textbf{E}ncoding (\textbf{FSCE}), a simple yet effective fine-tune based approach for few-shot object detection. When transfer the base detector to few-shot novel data, we augment the primary Region-of-Interest (RoI) head with a contrastive branch, the contrastive branch measures the similarity between object proposal encodings. A supervised contrastive objective with specific considerations for detection will be optimized to reduce the variance of object proposal embeddings from the same category, while pushing different-category instances away from each other. The proposed contrastive objective, contrastive proposal encoding (CPE) loss, is employed to the original classification and localization objective in a multi-task fashion. The end-to-end training of our proposed method is identical to vanilla Faster R-CNN. 

To our best knowledge, we are the first to bring contrastive learning into few-shot object detection. Our simple design sets the new state-of-the-art in any shot (1, 2, 3, 5, 10, and 30), with up to $+8.8\%$ on the standard PASCAL VOC benchmark and $+2.7\%$ on the challenging COCO benchmark.

\section{Related Work}

\textbf{Few-shot learning.} Few-shot learning aims to recognize new concepts given limited labeled examples. Meta-learning approaches aim at training a meta-model on episodes of individual tasks such that it can adapt to new tasks with few samples \cite{finn_model-agnostic_2017,ren_meta-learning_2018,sun_meta-transfer_nodate,khodadadeh_unsupervised_2019,ravi_optimization_2017,nichol_reptile_nodate,rusu_meta-learning_2018}, known as ``learning-to-learn''. Deep metric-learning based approaches emphasize learning good feature representation embeddings that facilitate downstream tasks. The most intuitive metrics including cosine similarity~\cite{cos_face,cos_softmax,chen_closer_2020,deng_arcface_2019}, euclidean distance to class center~\cite{snell_prototypical_2017}, and graph distances~\cite{garcia_few-shot_2018}. Interestingly, hallucinator-based methods solve the data deficiency via learning to generate fake-data~\cite{wang_low-shot_2018}. Existing few-shot learners are mostly developed in the context of classification. In comparison, few-shot detection is more challenging as it involves both classification and localization, yet under-researched.

\textbf{Few-shot object detection.} There are two lines of work addressing the challenging few-shot object detection (FSOD) problem. First, meta-learning based approaches devise a stage-wise and periodic meta-training paradigm to train a meta-learner to help knowledge transfer from base classes. Meta R-CNN \cite{yan_meta_2020} meta-learns channel-wise attention layer for remodeling the RoI head. MetaDet \cite{wang_meta-learning_2019} applies a weight prediction meta-model to dynamically transfer category-specific parameters from the base detector. FSIW \cite{xiao_few_shot_2020} improves upon Meta R-CNN and FSRW \cite{kang_few-shot_2019} by more complex feature aggregation and meta-training on a balanced dataset. With the balanced dataset introduced in TFA \cite{wang_frustratingly_2020}, fine-tune based detectors are rowing over meta-learning based methods in performance, MPSR~\cite{wu_multi-scale_2020} sets the current state-of-the-art by mitigating the scale scarcity in few-shot datasets, but its generalizability is limited because the positive refinement branch contains manual decisions. RepMet~\cite{karlinsky_repmet_2018} attaches an embedding sub-net in RoI head to model a posterior class distribution. It utilizes advanced tricks including OHEM~\cite{shrivastava_training_2016} and SoftNMS~\cite{bodla_soft-nms_2017} but fails to catch up with current SOTA. We criticize complex algorithms as they can easily overfit and exhibit poor test results in FSOD. Instead, our insight here is that the degeneration of average precision (AP) for novel categories mainly comes from misclassifying novel instances as confusable categories, and we resort to contrastive learning to learn discriminative object proposal representations without complexing the model.

\textbf{Contrastive learning}
The recent success of self-supervised models can be attributed to the renewed interest in exploring contrastive learning. \cite{hjelm_learning_2018,wu_unsupervised_2018,oord_representation_2019,chuang_debiased_2020,he_momentum_2020,chen_improved_2020,chen_simple_2020,chen_big_2020}. Optimizing the contrastive objectives \cite{oord_representation_2019,cos_face,deng_arcface_2019,supervised_contrastive_learning} simultaneously maximize the agreement between similar instances defined as positive pairs and encourage the difference among dissimilar instances or negative pairs. With contrastive learning, the algorithm learns to build representations that do not concentrate on pixel-level details, but encoding high-level features effective enough to distinguish different images \cite{chen_simple_2020,he_momentum_2020,chen_improved_2020,chen_big_2020}. Supervised contrastive learning~\cite{supervised_contrastive_learning} extends the batch contrastive approach to supervised setting, but for image classification. 

To our best knowledge, this work is the first to integrate supervised contrastive learning~\cite{sun2014deep,supervised_contrastive_learning} into few-shot object detection. The state-of-the-art few-shot detection performance in any shot and all benchmarks demonstrate the effectiveness of our proposed method.

\begin{figure*}[t]
\includegraphics[width=1\textwidth]{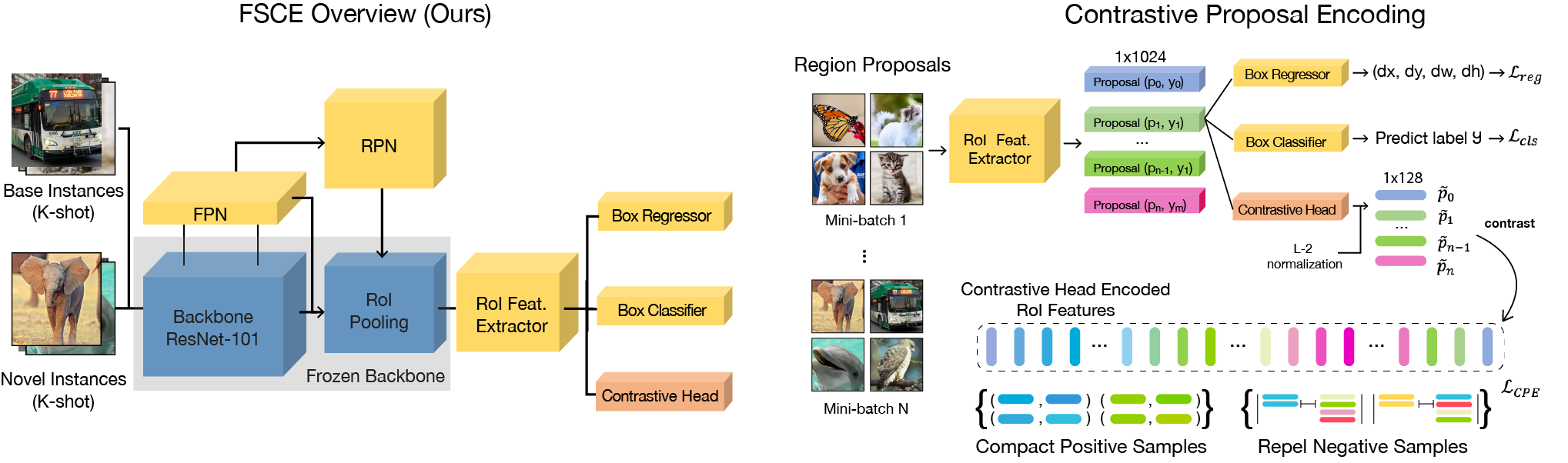}
\centering
\caption{Overview of our proposed FSCE. In our method, we jointly fine-tune the FPN pathway and RPN while fixing the backbone. We find this is effective in coordinating backbone feature maps to activate on novel objects yet still avoid the risk of overfitting. To learn contrastive object proposal encodings, we introduce a contrastive branch to guide the RoI features to learn contrastive-aware proposal embeddings. We design a contrastive objective to maximize the within-category agreement and cross-category disagreement.} 
\label{fig:overview}
\end{figure*}

\section{Method}
Our proposed method FSCE involves a simple two-stage training. First, the standard Faster R-CNN detection model is trained with abundant base-class data ($D_{train}=D_{base}$). Then, the base detector is transferred to novel data through fine-tuning on a balanced dataset~\cite{wang_cspnet_2019} with novel instances and randomly sampled base instances ($D_{train}=D_{novel}\cup D_{base}$). The backbone feature extractor is frozen during fine-tuning while the RoI feature extractor is supervised by a contrastive objective. We jointly optimize the contrastive proposal encoding (CPE) loss we proposed with the original classification and regression objectives in a multi-task fashion. Overview of our method is shown in Figure \ref{fig:overview}.

\subsection{Preliminary}

\textbf{Rethink the two-stage fine-tuning approach}. Original TFA~\cite{wang_frustratingly_2020} only fine-tunes the last two $fc$ layers--box classifier and box regressor--with novel data, the rest structures are frozen and taken as a fixed feature extractor. This could be viewed as an approach to counter the over-fitting of limited novel data. However it is counter-intuitive that Feature Pyramid Network (FPN~\cite{lin_feature_2017}), RPN, especially the RoI feature extractor which contain semantic information learned from base classes only, could be transferred directly to novel classes without any form of training.  In baseline TFA, unfreezing RPN and RoI feature extractor leads to degraded results for novel classes. However, we find this behavior is reversible and can  benefit novel detection results if trained properly. We propose a stronger baseline which adapts much better to novel data with jointly fine-tuned feature extractors and box predictors 

\begin{figure}[t]
\begin{center}
\includegraphics[width=1\linewidth]{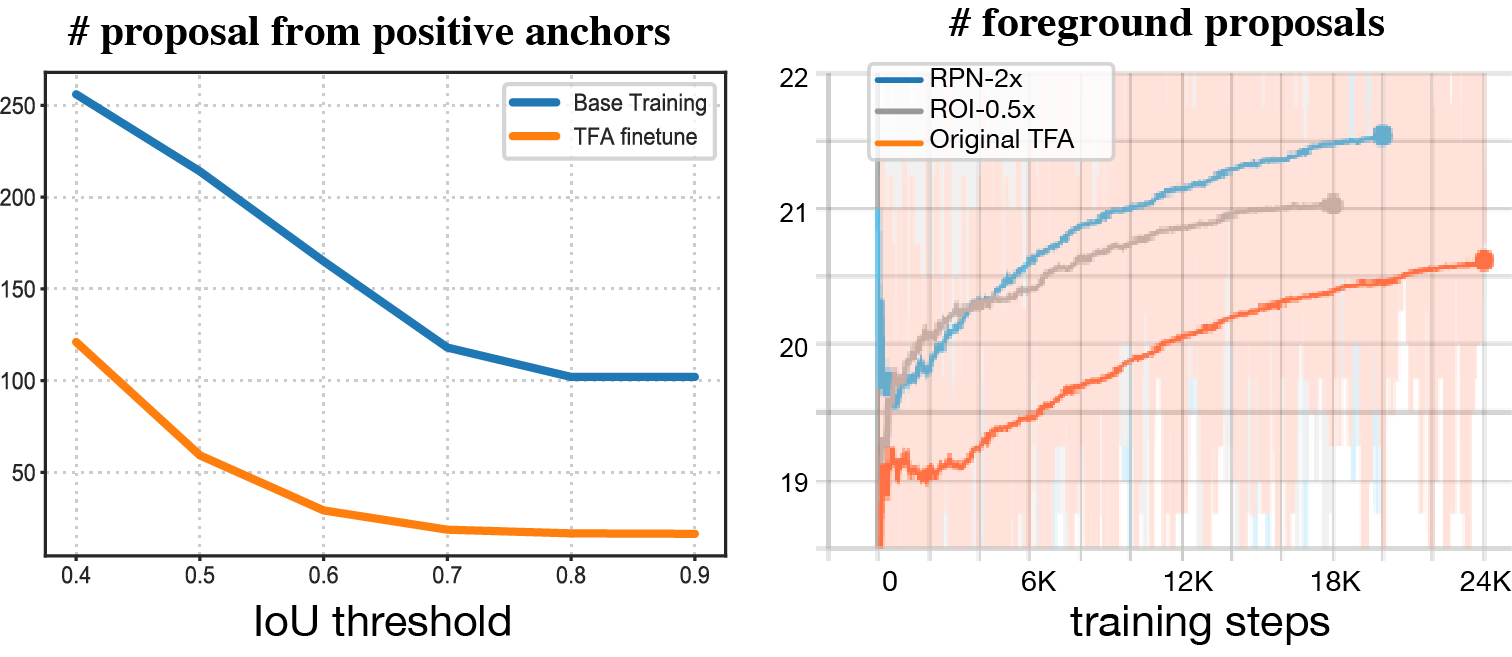}
\end{center}
\vspace{-2.5mm}
\caption{Key detection statistics. Left shows the average number of positive anchors per image in RPN in base training and novel fine-tuning stage. Right shows the average number of foreground proposals per image during fine-tuning. In the left, orange line shows the original TFA setting, which use the same specs as base training. In the right, the blue line shows double the number of anchors kept after NMS in RPN, the gray line shows reducing RoI head batch size by half.} 
\label{fig:rpn-roi}
\end{figure}

\textbf{Strong baseline.} We establish our strong baseline from the following observations. Initially, the detection performance for novel classes decreases as more network components are fine-tuned with novel shots. However, we notice a significant gap in the key RPN and RoI statistics between the data-abundant base training stage and the novel fine-tuning stage. As shown in Figure \ref{fig:rpn-roi}, the number proposals from positive anchors in novel fine-tuning is only $\frac14$ of its base training counterpart and the number of foreground proposals decreases consequently. We observe, especially at the beginning of fine-tuning, the positive anchors for novel objects receive comparatively low scores from RPN. Due to the low objectness scores, less positive anchors can pass non-max suppression (NMS) and become proposals that provide actual learning opportunities in RoI head for novel objects. Our insight is to rescue the low objectness positive anchors that are suppressed. Besides, re-balancing the foreground proposals fraction is also critical to prevent the diffusive yet easy backgrounds from dominating the gradient descent for novel instances in fine-tuning.

\begin{table}[t]\footnotesize
\vspace{1.5mm}
\begin{center}
\begin{tabular}{c|c|cc|cc}
\toprule
\multirow{2}{*}{Method} & Fine-tune & \multicolumn{2}{c|}{Refinement} & \multicolumn{2}{c}{Novel AP50}\\
 & FPN & RPN & ROI & ~~~~3 & 10 \\
\midrule
TFA w/ cos~\cite{wang_frustratingly_2020} & - & - & - & 44.7 & 56.0 \\
\midrule
\multirow{4}{*}{\shortstack{Strong baseline\\(Ours)}} & \cmark & \xmark & \xmark & 45.3 & 57.1 \\
&\cmark & \cmark & \xmark &  47.2 & 59.8 \\
&\cmark & \cmark & \cmark  & 49.7 & 61.4 \\
\bottomrule
\end{tabular}
\end{center}
\vspace{-1mm}
\caption{Novel detection performance of our strong baseline on PASCAL VOC Novel Split 1.}
\label{table:strongbaseline}
\end{table}

We use unfrozen RPN and ROI with two modifications, (1) double the maximum number of proposals kept after NMS, this brings more foreground proposals for novel instances, and (2) halving the number of sampled proposals in RoI head used for loss computation, as in fine-tuning stage the discarded half contains only backgrounds (standard RoI batch size is 512, and the number of foreground proposals are far less than half of it). As shown in Table \ref{table:strongbaseline}, our strong baseline boosts the baseline TFA by non-trivial margins. Moreover, the tunable RoI feature extractor opens up room for realizing our proposed contrastive object proposal encoding.

\subsection{Contrastive object proposal encoding}

In two-stage detection frameworks, RPN takes backbone feature maps as inputs and generates region proposals, RoI head then classifies each region proposal and regresses a bounding box if it is predicted to contain an object. In Faster R-CNN pipeline, RoI head feature extractor first pools the region proposals to fixed size and then encodes them as vector embeddings $x\in\mathbb{R}^{D_{R}}$ known as the RoI features. Typically $D_R=1024$ in Faster R-CNN w/ FPN. General detectors fail to establish robust feature representations for region proposals from limited shots, resulting in mislabeling localized objects and low average precision. The idea is to learn more discriminative object proposal embeddings, but according to our experiments, the category-level positive-margin classifier~\cite{cos_face,deng2009imagenet} does not work in this data-hungry setting. In order to learn more robust object feature representations from fewer shots, we propose to apply batch contrastive learning~\cite{supervised_contrastive_learning} to explicitly model instance-level intra-class similarity and inter-class distinction~\cite{sun2014deep,hadsell2006dimensionality} of object proposal embeddings. 

To incorporate contrastive representation learning into the Faster R-CNN framework, we introduce a contrastive branch to the primary RoI head, parallel to the classification and regression branches. The RoI feature vector $x$ contains post-ReLU~\cite{rectified_nodate} activations thus is truncated at zero, so the similarity between two proposals embeddings can not be measured directly. Therefore, the contrastive branch applies a 1-layer multi-layer-perceptron (MLP) head with negligible cost to encode the RoI feature to contrastive feature $z\in \mathbb{R}^{D_C}$, by default $D_C=128$. Subsequently, we measure similarity scores between object proposal representations on the MLP-head encoded RoI features and optimize a contrastive objective to maximize the agreement between object proposals from the same category and promote the distinctiveness of proposals from different categories. The proposed contrastive loss for object detection is described in the next section.

We adopt a cosine similarity based bounding box classifier, where the logit to predict $i$-th instance as $j$-th class is computed by the scaled cosine similarity between the RoI feature $x_i$ and the class weight $w_j$ in the hypersphere,
\vspace{-1.7mm}
\begin{equation}
	logit_{\{i,j\}} = \alpha\frac{x_i^\top w_j}{||x_i||\cdot||w_j||}
\end{equation}
$\alpha$ is a scaling factor to enlarge the gradient. We empirically fix $\alpha=20$ in our experiments. The proposed contrastive branch guides the RoI head to learn contrastive-aware object proposal embeddings which ease the discrimination between different categories. In the cosine projected hypersphere, our contrastive object proposal embeddings form tighter clusters with enlarged distances between different clusters, therefore increasing the generalizability of the detection model in the few-shot setting.

\begin{table*}[htbp]
\begin{center}
\resizebox{\textwidth}{!}{\begin{tabular}{l|c|ccccc|ccccc|ccccc}
\toprule
\multirow{2}{*}{Method / Shot} & \multirow{2}{*}{Backbone} & \multicolumn{5}{c|}{Novel Split 1} & \multicolumn{5}{c|}{Novel Split 2}& \multicolumn{5}{c}{Novel Split 3}\\
 & & 1 & 2 & 3 & 5 & 10 & 1 & 2 & 3 & 5 & 10 & 1 & 2 & 3 & 5 & 10\\
\midrule
LSTD~~~~~~~~~~~~~~\textit{AAAI 18} ~\cite{chen_lstd_2018} & VGG-16 & 8.2 & 1.0 & 12.4 & 29.1 & 38.5 & 11.4 & 3.8 & 5.0 & 15.7 & 31.0 & 12.6 & 8.5 & 15.0 & 27.3 & 36.3\\
\midrule
YOLOv2-ft~~~~~~\textit{ICCV19}~\cite{wang_meta-learning_2019} & \multirow{3}{*}{YOLO V2} & 6.6 & 10.7 & 12.5 & 24.8 & 38.6 & 12.5 & 4.2 & 11.6 & 16.1 & 33.9 & 13.0 & 15.9 & 15.0 & 32.2 & 38.4\\
$^\dagger$FSRW~~~~~~~~~~~~\textit{ICCV 19}~\cite{kang_few-shot_2019}  &  & 14.8 & 15.5 & 26.7 & 33.9 & 47.2 & 15.7 & 15.3 & 22.7 & 30.1 & 40.5 & 21.3 & 25.6 & 28.4 & 42.8 & 45.9\\  
$^\dagger$MetaDet~~~~~~~~\textit{ICCV 19}~\cite{wang_meta-learning_2019} &  & 17.1 & 19.1 & 28.9 & 35.0 & 48.8 & 18.2 & 20.6 & 25.9 & 30.6 & 41.5 & 20.1 & 22.3 & 27.9 & 41.9 & 42.9\\
\midrule
$^\ddagger$RepMet~~~~~~~~~\textit{CVPR 19}~\cite{karlinsky_repmet_2018} & InceptionV3& 26.1 & 32.9 & 34.4 & 38.6 & 41.3 & 17.2 & 22.1 & 23.4 & 28.3 & 35.8 & 27.5 & 31.1 & 31.5 & 34.4 & 37.2 \\

\midrule
FRCN-ft~~~~~~~~~\textit{ICCV 19} \cite{wang_meta-learning_2019} & \multirow{4}{*}{FRCN-R101} & 13.8 &  19.6 &  32.8 &  41.5 &  45.6 &  7.9 &  15.3 &  26.2 &  31.6 &  39.1 &  9.8 &  11.3 &  19.1 &  35.0 &  45.1\\
FRCN+FPN-ft~\textit{ICML}~20\cite{wang_frustratingly_2020} &&  8.2 &  20.3 &  29.0 &  40.1 &  45.5 &  13.4 &  20.6 &  28.6 &  32.4 &  38.8 &  19.6  & 20.8 &  28.7 &  42.2  & 42.1\\
$^\dagger$MetaDet~~~~~~~~\textit{ICCV 19}~\cite{wang_meta-learning_2019} &  &  18.9 &  20.6 &  30.2 &  36.8 &  49.6 &  21.8 &  23.1 &  27.8 &  31.7 &  43.0 &  20.6 &  23.9 &  29.4 &  43.9 &  44.1\\
$^\dagger$Meta R-CNN~\textit{ICCV 19}~\cite{yan_meta_2020} &  & 19.9 &  25.5 &  35.0 &  45.7 &  51.5 &  10.4  & 19.4 &  29.6 &  34.8  & 45.4  & 14.3  & 18.2  & 27.5 &  41.2 &  48.1\\
\midrule
TFA w/ fc~~~~~~~~\textit{ICML 20}~\cite{wang_frustratingly_2020} & \multirow{4}{*}{FRCN-R101} & 36.8 &  29.1 &  43.6 &  55.7 &  57.0 &  18.2 &  29.0 &  33.4 &  35.5 &  39.0 &  27.7 &  33.6 &  42.5 &  48.7 &  50.2\\
TFA w/ cos~~~~~~\textit{ICML 20} \cite{wang_frustratingly_2020} &  & 39.8 &  36.1 &  44.7 &  55.7 &  56.0 &  23.5 &  26.9 &  34.1 &  35.1 &  39.1 &  30.8 &  34.8 &  42.8 &  49.5 &  49.8\\
MPSR ~~~~~~~~~~~~\textit{ECCV 20}~\cite{wu_multi-scale_2020} & & 41.7 & - & 51.4 & 55.2 & 61.8 & 24.4 & - & 39.2 & 39.9 & 47.8 & 35.6 & - & 42.3 & 48.0 & 49.7\\
\rowcolor{mygray}
FSCE~~(Ours) & & \bf{44.2} & \bf{43.8} & \bf{51.4} & \bf{61.9} & \bf{63.4} & \bf{27.3} & \bf{29.5} & \bf{43.5} & \bf{44.2} & \bf{50.2} & \bf{37.2} & \bf{41.9} & \bf{47.5} & \bf{54.6} & \bf{58.5}\\

\bottomrule
\toprule
TFA w/ cos$^\star$~~~~\textit{ICML 20}~\cite{wang_frustratingly_2020} & \multirow{3}{*}{FRCN-R101} & 25.3 &  36.4 &  42.1 &  47.9 &  52.8 &  18.3 &  27.5 &  30.9 &  34.1 &  39.5 &  17.9 & 27.2 & 34.3 & 40.8 & 45.6\\
$^\dagger$FSIW$^\star$ ~~~~~~~~~~\textit{ECCV 20}~\cite{xiao_few_shot_2020} &  & 24.2 & 35.3 & 42..2 & 49.1 & 57.4 & 21.6 & 24.6 & 31.9 & 37.0 & 45.7 & 21.2 & 30.0 & 37.2 & 43.8 & 49.6\\
\rowcolor{mygray}
FSCE$^\star$~~(Ours) &  & \textbf{32.9} & \textbf{44.0} & \textbf{46.8} & \textbf{52.9} & \textbf{59.7} & \textbf{23.7} & \textbf{30.6} & \textbf{38.4} & \textbf{43.0} & \textbf{48.5} & \textbf{22.6} & \textbf{33.4} & \textbf{39.5} & \textbf{47.3} & \textbf{54.0} \\
\bottomrule
\end{tabular}}
\end{center}
\vspace{-1.7mm}
\caption{Performance evaluation (nAP 50) of existing few-shot detection methods on three PASCAL VOC Novel Split sets.
$\dagger$ marks meta-learning based methods. $\star$ represents average over 10 random seeds. $^\ddagger$ marks methods use \textit{N}-way \textit{K}-Shot meta-testing, which is a different evaluation protocol, see in Sec. \ref{sec41}.
}
\label{table:voc-main}
\end{table*}

\subsection{Contrastive Proposal Encoding (CPE) Loss}\label{lossfunc}
Inspired by supervised contrastive objectives in classification~\cite{supervised_contrastive_learning} and identification~\cite{sun2014deep}, our $CPE$ loss is defined as follows with considerations tailored for detection. Concretely, for a mini-batch of $N$ RoI box features $\{z_i, u_i, y_i\}_{i=1}^{N}$, where $z_i$ is contrastive head encoded RoI feature for $i$-th region proposal, $u_i$ denotes its Intersection-over-Union (IOU) score with matched ground truth bounding box, and $y_i$ denotes the label of the ground truth,
\vspace{1mm}
\begin{equation}
\mathcal{L}_{CPE} = \frac1N\sum_{i=1}^Nf(u_i)\cdot\mathit{L}_{z_i}  \label{loss3}
\end{equation}
\begin{equation}\small
    {\mathit{L}_{z_i}\! =\! \frac{-1}{N_{y_i}\!-1}\sum_{j=1,j\ne i}^{N}\mathbb{I}{\{y_i\!=\!y_j\}}\!\cdot\!\log\frac{\exp(\tilde{z_i}\!\cdot\! \tilde{z_j}/\tau)}{\sum_{k=1}^N\mathbb{I}_{k\ne i}\cdot\exp(\tilde{z_i}\!\cdot\! \tilde{z_k}/\tau)}}
    \label{loss4}
\end{equation}

 $N_{y_i}$ is the number of proposals with the same label as $y_i$, and $\tau$ is the hyper-parameter temperature as in InfoNCE~\cite{oord_representation_2019}.
 
In the above formula, $\tilde{z_i}$=$\frac{z_i}{||z_i||}$ denotes normalized features hence $\tilde{z_i}\cdot\tilde{z_j}$ measures the cosine similarity between the $i$-th and $j$-th proposal in the projected hypersphere. The optimization of the above loss function increases the instance-level similarity between object proposals with the same label and spaces proposals with different labels apart in the projection space. As a result, instances from each category will form a tighter cluster, and the margins around the periphery of the clusters are enlarged. The effectiveness of our $CPE$ loss has been confirmed by t-SNE visualization, as shown in Figure \ref{fig:tsne} (a) and (b).   

\textbf{Proposal consistency control.} Unlike image classification where semantic information comes from the entire image, classification signals in detection come from region proposals. We propose to use an IoU threshold to assure the consistency of proposals that are used to be contrasted, with the consideration that low IoU proposals deviate too much from the center of regressed objects, therefore might contain irrelevant semantics. In the formula above, $f(u_i)$ controls the consistency of proposals, defined with proposal consistency threshold $\phi$, and a re-weighting function $g(\cdot)$,
\begin{equation}
	f(u_i) = \mathbb{I}\{u_i\geqslant \phi\}\cdot g(u_i) \label{loss6}
\end{equation}
$g(\cdot)$ assigns different weight coefficients for object proposals with different level of IoU scores. We find $\phi$=0.7 is a good cut-off such that the contrastive head is trained with most centered object proposals. Ablations regarding $\phi$ and $g$ are shown in Sec. \ref{ablation}. 

\textbf{Training objectives.} In the first stage, the base detector is trained with a standard Faster R-CNN loss~\cite{ren_faster_2016}, a binary cross-entropy loss $\mathcal{L}_{rpn}$ to make foreground proposals from anchors, a cross-entropy loss $\mathcal{L}_{cls}$ for bounding box classifier, and a smoothed-$L1$ loss $\mathcal{L}_{reg}$ for box regression deltas. When transfer to novel data in the fine-tuning stage, we find the contrastive loss can be added to the primary Faster R-CNN loss in a multi-task fashion without destabilizing the training,
\begin{equation}
    \mathbb{L} = \mathcal{L}_{rpn} + \mathcal{L}_{cls} + \mathcal{L}_{reg} + \lambda\mathcal{L}_{CPE}
\end{equation}
$\lambda$ is set to 0.5 to balance the scale of the losses.

\begin{table}[t]\small
\begin{center}
\resizebox{1\linewidth}{!}{\begin{tabular}{ll|cc|cc}
\toprule
\multirow{2}{*}{Method} & \multirow{2}{*}{Year} & \multicolumn{2}{c|}{Novel AP} & \multicolumn{2}{c}{Novel AP75} \\
                        && 10 & 30 & 10 & 30 \\
\midrule
LSTD~\cite{chen_lstd_2018} & AAAI \textit{18} & 3.2 & 6.7 & - & - \\
$^\dagger$FSRW~\cite{kang_few-shot_2019} & ICCV \textit{19} & 5.6 & 9.1 & 4.6 & 7.6 \\
$^\dagger$MetaDet~\cite{wang_meta-learning_2019} & ICCV \textit{19} & 7.1 & 11.3 & 5.9 & 10.3 \\
$^\dagger$Meta-RCNN~\cite{yan_meta_2020} & ICCV \textit{19} & 8.7 & 12.4 & 6.6 & 10.8 \\
MPSR~\cite{wu_multi-scale_2020} & ECCV 20 & 9.8 & 14.1 & 9.7 & 14.2  \\
TFA w/ cos~\cite{wang_frustratingly_2020} & ICML \textit{20} & 10.0 & 13.7 & 9.3 & 13.4 \\
\rowcolor{mygray}
Ours & \multicolumn{1}{c|}{N/A} & \textbf{11.9} & \textbf{16.4} & \textbf{10.5} & \textbf{16.2} \\
\bottomrule
\toprule
TFA w/ cos$^\star$ ~\cite{wang_frustratingly_2020} & ICML \textit{20} & 9.1 & 12.1 & 8.8 & 12.0 \\
$^\dagger$FSIW$^\star$ \cite{xiao_few_shot_2020} & ECCV \textit{20} & \bf{12.5} & 14.7 & 9.8 & 12.2 \\
\rowcolor{mygray}
Ours$^\star$ & \multicolumn{1}{c|}{N/A} & 11.1 & \textbf{15.3} & \textbf{9.8} & \textbf{14.2} \\
\bottomrule
\end{tabular}}
\end{center}
\vspace{-1mm}
\caption{Few-shot detection evaluation results on COCO. $\star$ represents average over 10 random seeds. $\dagger$ marks meta-learning based methods.}
\label{table:coco-main}
\end{table}

\section{Experiments}\label{sec4}

Extensive experiments are performed in both PASCAL VOC~\cite{voc07, voc12} and COCO~\cite{coco} benchmarks. Our FSCE forms an upper envelope for all fine-tuning based methods and memory-inefficient meta-learns with large margins in any shots in all data splits. We strictly follow the consistent few-shot detection data construction and evaluation protocol~\cite{kang_few-shot_2019,wang_frustratingly_2020,wu_multi-scale_2020,xiao_few_shot_2020} to ensure fair and direct comparison. In this section, we first describe the few-shot detection settings, then provide complete comparisons of contemporary few-shot detection works on PASCAL VOC and COCO benchmarks, and provide ablation studies.

\textbf{Implementation Details.} For the detection model, we use Faster-RCNN~\cite{ren_faster_2016} with Resnet-101~\cite{he_deep_2015} and Feature Pyramid Network~\cite{lin_feature_2017}. All experiments are run on 8 GPUs with standard batch-size 16. The solver is standard SGD with momentum 0.9 and weight decay 1e-4. Naturally, we scale the training steps when training number of shots. Every detail will be open-sourced in a self-contained codebase to facilitate future research.

\subsection{Few-shot detection benchmarks}\label{sec41}
\vspace{-2.5mm}
\textbf{PASCAL VOC.} The overall 20 categories in PASCAL VOC are divided into 15 base categories and 5 novel categories. All base category data from PASCAL VOC 07+12 trainval sets are considered available, and \textit{K}-shot of novel instances are randomly sampled from previously unseen novel classes for $K=1,2,3,5$ and $10$. Following existing works \cite{wang_frustratingly_2020,kang_few-shot_2019,xiao_few_shot_2020}, we consider the same three random partitions of base and novel categories and samplings introduced in~\cite{kang_few-shot_2019}, referred as Novel Split 1, 2, and 3. And we report AP50 for novel predictions (nAP50) on PASCAL VOC 2007 test set. Note, this is different from the \textit{N}-Way \textit{K}-shot settings commonly used in meta-learning based methods~\cite{karlinsky_repmet_2018}. The huge variance between different random runs make the \textit{N}-Way \textit{K}-shot evaluation protocol unsuitable for few-shot object detection. For methods that provide results over 10 random seeds, we provide the corresponding results to compare with. 

\vspace{-2.5mm}
\textbf{MS COCO.} Similarly, for the 80 categories in COCO, 20 categories in common with PASCAL VOC are reserved as novel classes, the rest 60 categories are used as base classes. The $K=10$ and $30$ shots detection performance are evaluated on 5K images from COCO 2014 val dataset, COCO-style AP and AP75 for novel categories are reported by convention.

\vspace{-1.2mm}
\subsection{Few-shot detection results}
\vspace{-2mm}
\textbf{PASCAL VOC Results.} Results for all three random novel splits from PASCAL VOC are shown in Table \ref{table:voc-main}. Our FSCE outperforms all existing works in any shot and all splits. The effectiveness of our method is fully demonstrated. We are the first to achieve $>$50 nAP50 on split 2 and split 3, with up to +8.8 nAP50 above current SOTA on split 3. At the same time, our contrastive proposal encodings powered FSCE persists the less base forgetting property as in TFA. Demonstrated below in Table \ref{table:bf}.  

\vspace{-1.2mm}
\begin{table}[h]\normalsize
\begin{center}
\resizebox{\linewidth}{!}{\begin{tabular}{cccc|ccc}
\toprule
\multirow{2}{*}{Method} & \multicolumn{3}{c}{Base AP50} & \multicolumn{3}{c}{Novel AP50}\\
& 1 &3 &5 & 1  & 3 & 5\\
\midrule
Baseline-FPN~\cite{wu_multi-scale_2020} & 56.9 & 66.2 & 67.9 & 25.5 & 41.1 & 49.6\\
MPSR~\cite{wu_multi-scale_2020} & 59.4 & 67.8 & 68.4 & 41.7 & 51.4 & 55.2\\
TFA w/ cos (Our impl.)& \bf{79.1} & \bf{77.3} & \bf{77.0} & 39.8 & 44.6 & 55.6 \\
\rowcolor{mygray}
FSCE (Ours) & 78.9 & 74.1 & 76.6 & \bf{44.2} & \bf{51.4}& \bf{61.9} \\
\bottomrule
\end{tabular}}
\end{center}
\vspace{-2mm}
\caption{Base forgetting comparisons on PASCAL VOC Split 1. Before fine-tuning, the base AP50 in base training is 80.8. }
\label{table:bf}
\end{table}

\begin{figure*}[h]
\begin{center}
\includegraphics[width=1\textwidth]{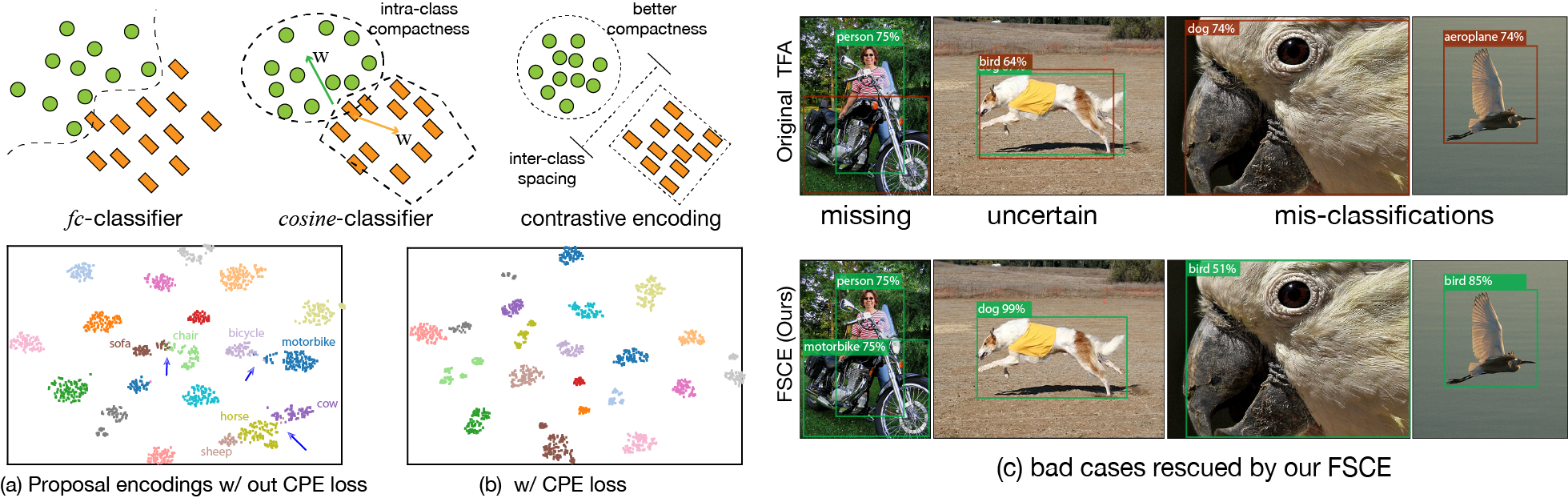}
\end{center}
\vspace{-2mm}
\caption{Conceptually and t-SNE visualization of the object proposal embeddings learned with and without our $CPE$ loss, our $CPE$ loss explicitly model the within-class similarity and cross-class distance. t-SNE here shows the proposal encodings from randomly selected 200 PASCAL VOC images. Right panel shows bad cases rescued by our contrastive-aware representations.} 
\label{fig:tsne}
\end{figure*}

\vspace{-2.5mm}
\textbf{COCO Results.} Few-shot detection results for COCO are shown in Table \ref{table:coco-main}. Our FSCE set new state-of-the-art for all shots, under the same testing protocol and same metrics. Our proposed methods gain +1.7 nAP and +2.7 nAP75 above current SOTA, which is more significant than the gaps between any previous advancements.

\subsection{Ablation}\label{ablation}

\textbf{Components of our proposed FSCE.} First, with our modified training specification for fine-tune stage, the class-agnostic RPN and RoI head can be directly transferred to novel data and incur huge performance gain, this is because we utilize more low-quality RPN proposals that would normally be suppressed by NMS and provide more foregrounds to learn given the limited optimization opportunity in few-shot setting. And the jointly fine-tuned FPN top-down convolution and RoI feature extractor opens up room for better representation learning. Second, our $CPE$ loss guides the RoI feature extractor to establish contrastive-aware objects embeddings, intra-class compactness and inter-class variance ease the classification task and rescue misclassifications. The whole system benefits from the proposal consistency control by employing only high-IoU region proposals that are less deviated from objects center to contrast. All ablation studies are done with PASCAL VOC Novel Split 1 unless otherwise specified.

\begin{table}[h]\Large
\begin{center}
\resizebox{\linewidth}{!}{\begin{tabular}{ccccc|ccc}
\toprule
\multirow{2}{*}{Method} & \multicolumn{2}{c}{Refinement} & \multirow{2}{*}{\shortstack{CPE\\loss}} & \multirow{2}{*}{\shortstack{Proposal\\Consistency}} & \multicolumn{3}{c}{Novel AP50}\\
\cline{2-3}
 & RPN & ROI & & & 3 & 5 & 10 \\
\midrule
TFA w/ cos & \xmark & \xmark & - & - & 44.7 & 55.7 & 56.0\\
\midrule
\multirow{4}{*}{\shortstack{FSCE\\(Ours)}} & \cmark & \xmark & \xmark & \xmark & 47.2 & 56.9 & 59.8\\
& \cmark & \cmark & \xmark  & \xmark & 49.7 & 58.6 & 61.4 \\
& \cmark & \cmark & \cmark & \xmark & 50.6 & 60.7 & 62.7 \\
& \cmark & \cmark & \cmark & \cmark & \bf{51.4} & \bf{61.9} & \bf{63.4} \\
\bottomrule
\end{tabular}}
\end{center}
\caption{Ablation for key components proposed in FSCE.}
\end{table}

\textbf{Ablation for contrastive branch hyper-parameters.} Primary RoI feature vector contains post-ReLU activations truncated at zero, we therefore encode the RoI feature with a contrastive head to $z\in \mathbb{R}^{D_C}$ such that similarity can be meaningfully measured. Based on our ablations, the few-shot detection performance is not sensitive to the contrastive head dimension. And among the commonly used temperature $\tau$ used in contrastive objectives~\cite{supervised_contrastive_learning,he_momentum_2020,chen_improved_2020}, a medium temperature $\tau=0.2$ works better than relatively small value 0.07 and large value 0.5.
\begin{table}[h]\small
\begin{center}
\begin{tabular}{c|ccc}
\toprule
\multirow{2}{*}{\shortstack{Contrast Head\\Dimension}} & \multicolumn{3}{c}{Temperature~($\tau$)} \\
& 0.07 & 0.2 & 0.5 \\
\midrule
$D_C=128$& 63.1 & \bf{63.4} & 62.9 \\
\midrule
$D_C=256$& 62.4 & \bf{63.4} & 63.3\\
\bottomrule
\end{tabular}
\end{center}
\vspace{-1.5mm}
\caption{Ablation for contrastive hyper-parameters, results from 10 shot of PASCAL VOC Split 1.}
\end{table}

\textbf{Ablation for Proposal Consistency Control.}
In equation (\ref{loss4}) and (\ref{loss6}), we propose a compound proposal consistency control mechanism, comprised of an indicator function with an IoU cut-off threshold $\phi$, and a function $g(\cdot)$ for re-weighting proposals with different level of IoU. Turns out a re-weighting is not necessary and a simple high-IoU cut-off works the best for 5 and 10 shots, but when number of shots is low, simply filtering out proposals with IoU less than $\phi$ becomes less favorable as the data sparsity is too severe. In low-shot cases, keeping all proposals but down-weight low-IoU ones make more sense, and empirically, exponential decay (easy mining) does worse than a simple linear weighting.

\begin{table}[h]\normalsize
\label{ab-loss}
\begin{center}
\resizebox{\linewidth}{!}{\begin{tabular}{c|c|c|ccc}
\toprule
\multirow{2}{*}{Option} & \multirow{2}{*}{Threshold} & \multirow{2}{*}{\shortstack{Reweight\\function}} & \multicolumn{3}{c}{Novel AP} \\
& & & 3 & 5 & 10 \\
\midrule
\multirow{2}{*}{Hard Clip} & $\phi=0.5$ & $g(x)=1$ & 50.5 & 60.7 & 62.1\\
& $\phi=0.7$ & $g(x)=1$ & 50.8 & \bf{61.9} & \bf{63.4}\\
\midrule
\multirow{2}{*}{Weighting} & $\phi=0$ & $g(x)=x$&\bf{51.4}& 59.7 & 61.1 \\
 & $\phi=0$ & $g(x)=e^x-1$ & 50.8 & 59.6 & 61.6\\
\bottomrule
\end{tabular}}
\end{center}
\vspace{-1.5mm}
\caption{Ablation for proposal consistency control in FSCE.}
\end{table}

\textbf{Visual inspections and analysis.} Figure \ref{fig:tsne} shows visual inspections of our proposed FSCE. We find in data-abundant general detection, the saturated performance of $fc$ classifier and $cosine$ classifier are essentially equal. $fc$ layer can learn sophisticated decision boundary from enough data. Existing literature and we all confirm that $cosine$ box classifier excels in few shot object detection, this can be attributed to the explicitly modeled similarity helps form tighter instances clusters on the projected unit hypersphere. The intuition to spacing different categories is trivial, but per our experiments well-established margin-based classifiers~\cite{cos_face,deng_arcface_2019} does not work in this data-hunger setting (-2 nAP compared to FSCE in 10 shots and worse in lower shots). Instead of adding a margin to classifier, FSCE models the instance-level intra-class similarity and inter-class via $CPE$ loss and guide RoI head to learn contrastive-aware object proposal representations. t-SNE~\cite{t_sne} visualization of objects proposal embeddings affirms the effectiveness of our $CPE$ loss in reducing intra-class variance and form more defined decision boundaries, this aligns well with our proposition. Figure \ref{fig:tsne} (c) shows example bad cases from TFA that are rescued by our FSCE including, missing detection for novel instances, low confidence scores for novel instances, and the pervasive misclassifications.

\vspace{-2mm}
\section{Conclusion}
\vspace{-2mm}
In this work, we propose a new perspective of solving FSOD via contrastive proposals encoding. Effectively saving accurately localized objects from being misclassified, our method achieves state-of-the-art results in any shot and both benchmarks, with up to +$8.8$\% on PASCAL VOC and +$2.7$\% on COCO. Our proposed contrastive proposal encoding head has a negligible cost and is generally applicable. It can be chipped into any two-stage detectors without interfering with the training pipeline. Also, we provide a strong baseline comparable to contemporary SOTA to facilitate future research in FSOD. 
For a broader impact, FSOD is of great worth considering the vast amount of objects in the real world. Our work proves the plausibility of incorporating contrastive learning into object detection frameworks. We hope our work can inspire more researches in contrastive visual embedding and few-shot object detection.

\textbf{Acknowledgement.} This work was supported by grants from the National Key R \& D Program of China. Grant number: 52019YFB1600500.

{\small
\bibliographystyle{unsrt}
\bibliographystyle{ieee_fullname}
\bibliography{cvpr_final.bib}
}

\end{document}